\documentclass{jdmdh}
\usepackage{array}
\usepackage{pgfplots}
\usepackage{placeins}
\pgfplotsset{compat=1.18}
\usepackage{subcaption}
\usepackage{multirow}
\usepackage{graphicx}

\titlespacing*{\section}
{0pt}{2.5ex plus 1ex minus .2ex}{1.3ex plus .2ex}
\titlespacing*{\subsection}
{0pt}{1.0ex plus 1ex minus .2ex}{1.3ex plus .2ex}
\titlespacing*{\subsubsection}
{0pt}{1.0ex plus 1ex minus .2ex}{1.3ex plus .2ex}

\title{Incorporating Crowdsourced Annotator Distributions into Ensemble Modeling to Improve Classification Trustworthiness for Ancient Greek Papyri}
\author[1]{Graham West}
\author[1]{Matthew I. Swindall}
\author[4]{Ben Keener}
\author[4]{Timothy Player}
\author[3]{\\Alex C. Williams}
\author[2]{James H. Brusuelas}
\author[1]{John F. Wallin}

\affil[1]{Middle Tennessee State University, USA} 
\affil[2]{University of Kentucky, USA}
\affil[3]{Amazon, USA}
\affil[4]{University of Tennessee, Knoxville, USA} 

\corrauthor{Matthew I. Swindall}{mis2n@mtmail.mtsu.edu}

\begin{document}

\maketitle

\abstract{Performing classification on noisy, crowdsourced image datasets can prove challenging even for the best neural networks. Two issues which complicate the problem on such datasets are class imbalance and ground-truth uncertainty in labeling. The AL-ALL and AL-PUB datasets---consisting of tightly cropped, individual characters from images of ancient Greek papyri---are strongly affected by both issues. The application of ensemble modeling to such datasets can help identify images where the ground-truth is questionable and quantify the trustworthiness of those samples. As such, we apply stacked generalization consisting of nearly identical ResNets with different loss functions: one utilizing sparse cross-entropy (CXE) and the other Kullback-Liebler Divergence (KLD). Both networks use labels drawn from a crowdsourced consensus. This consensus is derived from a Normalized Distribution of Annotations (NDA) based on all annotations for a given character in the dataset. For the second network, the KLD is calculated with respect to the NDA. For our ensemble model, we apply a $k$-nearest neighbors model to the outputs of the CXE and KLD networks. Individually, the ResNet models have approximately 93\% accuracy, while the ensemble model achieves an accuracy of $>$95\%, increasing the classification trustworthiness. We also perform an analysis of the Shannon entropy of the various models’ output distributions to measure classification uncertainty. Our results suggest that entropy is useful for predicting model misclassifications.}


\section{Introduction}

One of the greatest challenges in the domain of machine learning is the acquisition of useful datasets. Three main-stream approaches include:

\begin{itemize}
    \item Professional annotation services
    \item Synthetic dataset generation
    \item Crowdsourcing initiatives
\end{itemize}

Professional annotation services such as Amazon Mechanical Turk outsource labeling to human annotators who are unlikely to be domain experts, i.e. \citet{turk}. The approach can also be expensive and take a long time depending on the size of the dataset. While Generative Adversarial Networks (GANs) can produce astonishingly realistic synthetic image datasets, as in \citet{Swindall2022}, it is impractical to verify that every feature and label are correct for large scale projects. Crowdsourcing initiatives face some of the same challenges as the first approach, but with some additional ones. Due to the number of annotators, crowdsourcing makes it more practical to have multiple annotators review and annotate the same image. This can be a helpful mechanism for managing annotation errors, but it also leads to an inherent uncertainty in labeling. If multiple annotators disagree about the class to which an image belongs, what should its label be? Regardless of the approach used, there is usually a lack of certainty, i.e. "noise," in the ground-truth of the produced dataset.

In the field of papyrology (the study of ancient Greek papyri), these challenges are compounded by the lack of quality image datasets and a shortage of experts who are qualified to annotate them. Efforts to produce a crowdsourced image dataset from such papyri, such as AL-ALL and AL-PUB in \citet{Swindall2021} and \citet{williams2014}, have shown that, even though the ground-truth in labeling may be questionable, the utilization of annotator consensus can produce datasets that can be accurately modeled using deep learning approaches.

To build on these efforts, we propose a stacked generalization method comprised of two ResNets trained on the AL-ALL dataset with identical architectures but different loss functions and labeling schemes. It is important to clarify that the focus of this work, as well as our contribution, is the incorporation of this novel labeling scheme, not ensemble approaches in general. We thoroughly discuss this scheme, called Normalized Distribution of Annotations (NDA), in later sections. We then explore the utility of the Shannon entropy of the various models' output probability distributions as a measure of classification uncertainty. Our results demonstrate that the proposed ensemble method yields greater accuracy than either of the individual models for this dataset. It also highlights the ability of the Shannon entropy to quantify the uncertainties of both the model output distributions as well as the distribution of crowdsourced annotations.  

\subsection{Deep Learning and Handwritten Character Recognition}

While deep learning models can classify characters with astounding accuracy, ancient handwritten manuscripts still pose many challenges. Existing character recognition engines have been mostly trained on modern, printed documents and perform poorly for handwritten texts. In the case of highly damaged, ancient Greek papyri, the failure of such engines demonstrates the need for custom-trained models as in \citet{Swindall2021}.

\subsection{The Ancient Lives Project}

The Ancient Lives Project, detailed in \citet{williams2014}, was a web-based crowdsourcing initiative that allowed volunteers to transcribe digital images of ancient papyrus fragments from 2011 until 2018. These manuscripts consist of handwriting on papyrus, in various states of preservation, excavated from rubbish dumps near the ancient Egyptian city of Oxyrhynchus, as explained in \citet{OxyBook}. This collection of papyri, housed at the University of Oxford, continues to be a source of new discoveries for both Greek literature and the history of Roman Egypt. The goal of Ancient Lives was to create a dataset that would open up new avenues to studying this vast collection. Although the project began over a decade ago, only more recent advancements in machine learning, or artificial intelligence approaches, are making that original goal possible. This project resulted in millions of annotations from images of papyrus fragments, such as the one shown in Figure \ref{fig:Oxy}. A consensus algorithm was applied to these annotations to produce approximate locations of individual characters within the document images, as well as a consensus labels for each identified character. 

\subsection{Datasets: AL-ALL and AL-PUB}
\begin{figure}
    \begin{minipage}[c]{0.45\linewidth}
        \centering
        \includegraphics[width=0.5\linewidth]{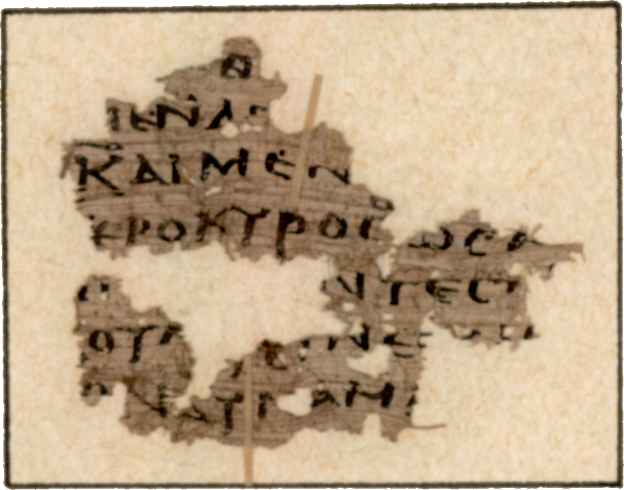}
            \caption{An Example of an Oxyrhynchus Papyrus: Fragment. Note the extensively damaged condition of the manuscript.}
            \label{fig:Oxy}
    \end{minipage}
    \hfill
    \begin{minipage}[c]{0.45\linewidth}
        \centering
        \includegraphics[width=0.5\linewidth]{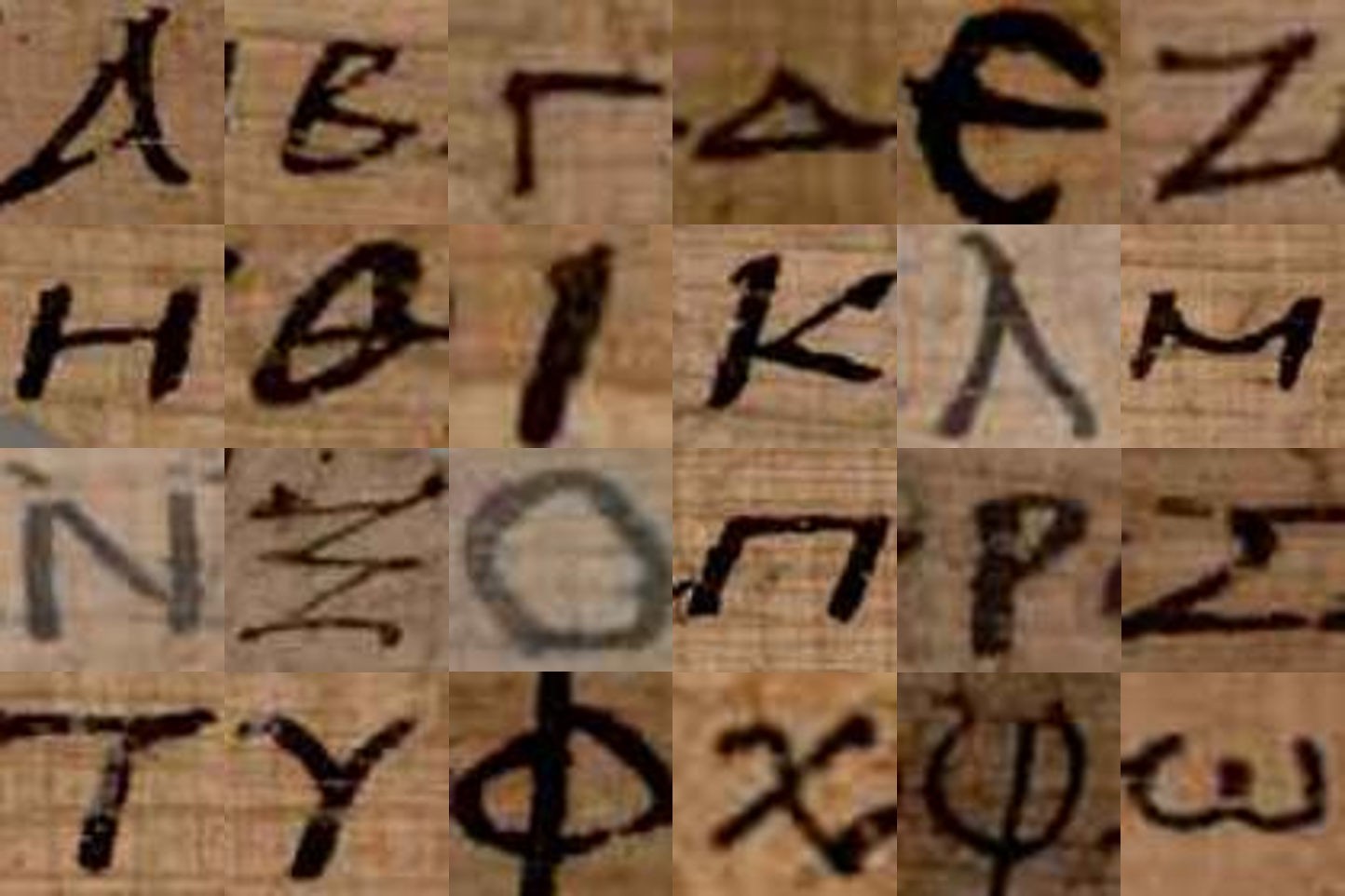}
            \caption{Examples of Character Images from the AL-PUB Dataset}
            \label{fig:alphabet}
    \end{minipage}
\end{figure}

The consensus data from the Ancient Lives Project was used to create two crowdsourced datasets: AL-ALL (containing 399,421 images from published and unpublished papyri) and AL-PUB (containing 195,683 images from only published papyri). Both datasets consist of tightly cropped images of individual Greek characters from the annotated papyri images. Figure \ref{fig:alphabet} shows example images from AL-PUB, one for each character in the 
standard Greek alphabet. Table \ref{tab:al_counts} lists the sample sizes for each character in AL-ALL. As some of the fragment images used to create AL-ALL are not yet in the public domain, AL-PUB was created to share with the public. AL-PUB is freely available for download at \textit{https://data.cs.mtsu.edu/al-pub/}. Though AL-PUB is a smaller subset of AL-ALL, \citet{Swindall2021} suggests minimal model performance impairment when training with AL-PUB versus AL-ALL. However, to maximize the benefits of the larger dataset, our image classification models were trained on AL-ALL. It is important to note that these datasets are drastically different from typical character image datasets such as MNIST, detailed in \citet{deng2012mnist}, which consists of a custom created dataset that is carefully curated with little-to-no uncertainty in the ground-truth labels. AL-ALL and AL-PUB are derived from heavily damaged papyrus fragments, excavated from ancient rubbish dumps. These papyri have largely been annotated by untrained citizen scientists and a small number of professional experts. Such a process introduces various and complex forms of noise into the dataset. 
\FloatBarrier

\begin{table}[t]
\tiny
\begin{minipage}[c]{0.45\linewidth}
\centering
\begin{tabular}{|l|c||l|c|}
    \hline
    \textbf{Character} & \textbf{Count} & \textbf{Character} & \textbf{Count}\\
    \hline
    \hline
    \textbf{Alpha} ($A, \alpha$) & 42,546 & \textbf{Nu} ($N, \nu$) & 44,910\\ \hline
    \textbf{Beta} ($B,\beta$) & 2,534 & \textbf{Xi} ($\Xi, \xi$) & 1,201 \\\hline
    \textbf{Gamma} ($\Gamma, \gamma$)& 6,907 & \textbf{Omicron} ($O, o$) & 46,344\\ \hline
    \textbf{Delta} ($\Delta, \delta$) & 11,717 & \textbf{Pi} ($\Pi, \pi$) & 17,114\\ \hline
    \textbf{Epsilon} ($E, \epsilon$) & 31,584 & \textbf{Rho} ($P, \rho$) & 20,450\\\hline
    \textbf{Zeta} ($Z, \zeta$) & 1,425 & \textbf{Sigma} ($\Sigma, \sigma$) & 62 \\\hline
    \textbf{Eta } ($H, \eta$) & 15,064 & \textbf{Tau} ($T, \tau$) & 32,045 \\\hline
    \textbf{Theta} ($\Theta, \theta$) & 7,575 & \textbf{Upsilon} ($Y, \upsilon$) & 15,762\\\hline
    \textbf{Iota} ($I, \iota$) & 25,595 & \textbf{Phi} ($\Phi, \phi$) & 6,063\\\hline
    \textbf{Kappa} ($K, \kappa$) & 17,937 & \textbf{Chi} (X, $\chi$) & 9,156\\\hline
    \textbf{Lambda} ($\Lambda, \lambda$) & 13,253 & \textbf{Psi} ($\Psi, \psi$) & 904 \\\hline
    \textbf{Mu} ($M, \mu$) & 13,227 & \textbf{Omega} ($\Omega, \omega$) & 16,046 \\ \hline
\end{tabular}
\caption{Counts for each letter in the Ancient Lives dataset.}\vspace{2mm}
\label{tab:al_counts}
\end{minipage}
\hfill
\begin{minipage}[c]{0.45\linewidth}
    \centering
        \includegraphics[width=\linewidth]{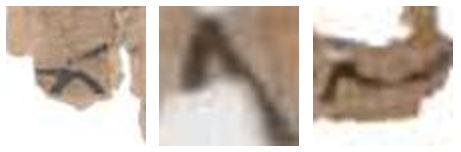}
            \captionof{figure}{Characters From Damaged Papyri}
            \label{fig:damaged}
\end{minipage}
\end{table}

\subsection{Sources of Noise in the Dataset}
The most prominent sources of noise in AL-ALL and AL-PUB can be split into three main categories that will be discussed below.
    
\subsubsection{Class Imbalance}
There is significant class imbalance inherent to the datasets due to multiple factors. The primary source of class imbalance is the lack of instances of certain characters in the source manuscripts themselves. This imbalance is compounded by the tendency of human annotators to ``chase characters,'' i.e., to preferentially annotate specific characters. While most characters have many thousands of samples (see Table \ref{tab:al_counts}), there are several which have a relatively small amount. The most significant of these imbalanced outliers is Sigma ($\Sigma$ $\sigma$), with only 62 samples. Due to its small sample size, Sigma is the most challenging character to classify accurately and our models perform very poorly on this character.

\subsubsection{Image Noise}

Unlike datasets such as MNIST, from \citet{deng2012mnist}, AL-ALL and AL-PUB were created from images containing holes, rips, missing segments, stains, and faded ink. As shown in Figure \ref{fig:damaged}, many character images are illegible, incomplete, or contain multiple characters. A further source of noise is the cropping algorithm used to create images of the individual characters. The cropping bounds are determined by approximating a manuscript's mean inter-character distance which is calculated from the character locations as determined by the annotators. Since characters on a manuscript are rarely evenly-spaced and the annotator locations are imperfect, this approach can lead to image crops which are zoomed too far out (including multiple characters) or too far in (eliminating portions of the character of interest).

\subsubsection{Annotator Noise}
In addition to the uncertainty in character locations, annotators supply an additional source of noise since they, at times, classify characters incorrectly and regularly disagree with other annotators. As most annotators are not trained professionals, they may easily confuse characters with similar shapes such as Alpha, Delta, and Lambda ($A, \Delta, \Lambda$). It is also possible for annotators to confuse a sequence of characters for a single character due to the connection (or near-connection) of their ligatures. For example: confusing two side-by-side Tau's ($\tau \tau$) for Pi ($\pi$).

\subsection{Normalized Distribution of Annotations (NDA)}
The amalgamation of all annotations for a single character in AL-ALL and AL-PUB can be treated as a sparse distribution over the set of 24 character classes. When normalized, this distribution is similar to a probability distribution. We illustrate this Normalized Distribution of Annotations (NDA) in Equation \ref{feature_scaling}. If we let $M=24$ classes and $x_i$ be the number of human annotations for class $i$ for some image (i.e., the values of the non-normalized distribution of annotations), then we can represent NDA of the image by $x_i'$ where
\begin{equation}
	x_i' = \frac{x_i}{\sum_{j=1}^{M} x_j}
	\label{feature_scaling}
\end{equation}

\begin{figure}
    \begin{minipage}[c]{0.45\linewidth}
        \centering
        \includegraphics[width=\linewidth]{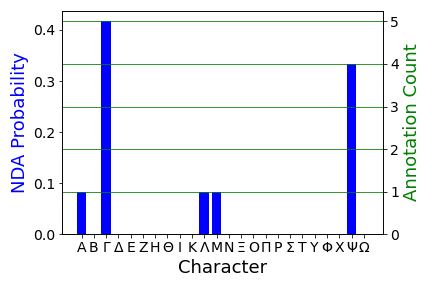}
        \caption{Normalized Distribution of Annotations (NDA) for the image Shown in Figure \ref{fig:gamma}.}
        \label{fig:nda_bar}
    \end{minipage}
    \hfill
    \begin{minipage}[c]{0.45\linewidth}
        \centering
        \includegraphics[width=0.3\linewidth]{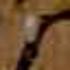}
            \caption{AL-PUB Gamma ($\Gamma \gamma$) Example Z\_POxy.v0015.n1805.a.01\_135790\_216\_Gamma\_5-12.jpg.}
            \label{fig:gamma}
    \end{minipage}   
\end{figure}

Figure \ref{fig:nda_bar} shows a single example of this NDA. This distribution is an excellent illustration of the annotator noise inherent in the datasets as the ground-truth is not crystal-clear. The character image corresponding to this NDA is shown in Figure \ref{fig:gamma} with a consensus label of Gamma ($\Gamma$). However, a large percentage of annotators (roughly 33\%) believed the character to be a Psi ($\Psi$). The extant damage in the papyrus fragment makes it challenging even for expert papyrologists to agree on the correct classification for this image. The NDA is reminiscent of, yet very different from, a non-continuous version of a Softmax output distribution in neural networks. In the Methods section, we make use of the NDA in the KLD loss function of our second ResNet.

\section{Methods}

In this section, we discuss the various methods used in this study. We describe the architecture of the ResNets, their labeling schemes, and their loss functions. We then describe our approach for performing stacked generalization to combine the ResNets' outputs to create input features for a $k$-nearest neighbors model. We discuss the evaluation metrics which were used to monitor the training progress of the ResNets. Finally, we discuss the use of Shannon entropy as a means to measure classification uncertainty.

An important semantic distinction must be addressed before further discussion. As the ground-truth can often be ambiguous for crowdsourced datasets, it is necessary to assume that the labels (derived from consensus of human annotations) are the ground-truth. Thus, when we indicate that a particular image was ``correctly classified'' by a given model, we are stating that the model prediction agrees with the annotation consensus. Conversely, when we state an image is ``incorrectly classified,'' we are indicating that the model disagrees with the consensus.

\subsection{Deep Learning Architecture}

The core of the proposed ensemble model consists of two, nearly identical architectures. The ResNet, outlined in Figure \ref{fig:resnet_arch}, is a fairly standard convolutional residual network, as in \citet{He_2016_CVPR}. The early layers in the model consist of two convolution layers with ReLU activation, followed by a max-pooling layer and 16 residual blocks. The residual blocks consist of two cycles of convolution and batch normalization. After the residual block, the model finishes with a global average pooling layer, a single dense layer, a dropout layer, and finally an output layer. These models differ only in the class labels and loss functions. The architecture is based on the models utilized in \citet{Swindall2021}. For both models, the input features are identical: the raw images. The first model, CXE-ResNet, takes standard, single value numerical class labels while the second model, KLD-ResNet, takes the NDA from the Ancient Lives Project data as labels. 

The ResNets, written in Tensorflow 2.3 with GPU acceleration, were trained on a Linux system consisting of 2 Intel Xeon E5-2687W Processors and 2 NVIDIA GeForce RTX 2080 Ti CUDA cores. In the interest of limiting the differences between the two ResNets, the same randomization seed was used for both models, though this likely had minimal effect on training or inference. Recent work, such as \citet{GPU_dominates}, suggests that GPU non-determinism dominates the effects of seed randomization. 

\begin{table}
    \begin{minipage}[c]{0.45\linewidth}
	   \centering
          \footnotesize
	   \begin{tabular}{|| c | c | c | c ||}
		  \hline
		  \multicolumn{2}{|| c |}{} & \multicolumn{2}{c||}{KLD} \\
		  \cline{3-4}
		  \multicolumn{2}{|| c |}{} & Correct & Incorrect  \\
		  \hline
		  \multirow{2}{2em}{CXE} & Corrrect & 353549 & 16407 \\
		  \cline{2-4}
		  & Incorrect & 18594 & 10871 \\
		  \hline
	   \hline
	   \end{tabular}
	       \caption{Number of images correctly/incorrectly classified by the CXE- and KLD-     ResNets. Both models agreed with consensus annotation for 353,549 images, and       both disagreed for 10,871 images.}
	       \label{confCK}
    \end{minipage}
    \hfill
    \begin{minipage}[c]{0.45\linewidth}
        \centering
        \includegraphics[width=0.75\linewidth]{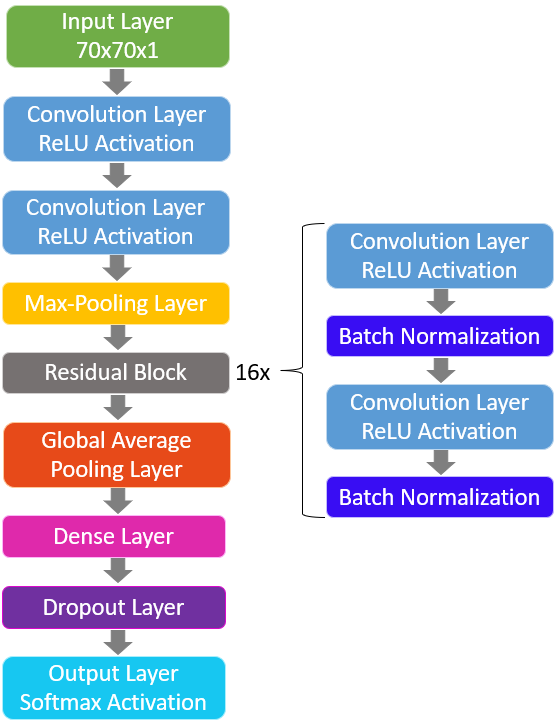}
        \captionof{figure}{ResNet Architecture}
        \label{fig:resnet_arch}
    \end{minipage}
\end{table}

\subsubsection{Loss functions}

Our ResNets used two different loss functions, respectively: sparse categorical cross-entropy (CXE) and Kullback-Leibler divergence (KLD). Both of these functions are based on the Shannon entropy $H$ of a probability distribution $p$ \citep{Shannon} (see Equation \ref{ent}).
\begin{equation}
	H(p) = \textbf{E}[ -\mathrm{log}(p) ] =  - \sum_{i=1}^{M} p_i \mathrm{log}(p_i)
	\label{ent}
\end{equation}
Here, the distribution $p=(p_1,\cdots,p_{M})$ is defined over $M$ classes. The entropy function (which takes the entire probability distribution as input) returns the amount of ``uncertainty'' present in the distribution. Thus, for distributions which are very narrowly peaked, the entropy will be low, while for distributions which are more spread out and uniform, the entropy will be high. The full range of values of the Shannon entropy is from $0$ to $\mathrm{log}(M)$.

The first of our loss functions---sparse categorical cross entropy---can be defined in terms of the Shannon entropy of $p$ \textit{with respect to} $q$ as in Equation \ref{wrt}
\begin{equation}
	H_q(p) = \textbf{E}_q[ -\mathrm{log}(p_i) ] = - \sum_{i=1}^{M} q_i \mathrm{log}(p_i)
	\label{wrt}
\end{equation}
where $p$ is the model's output distribution and $q$ the ``target'' distribution. Since we are using \emph{sparse} CXE, this means that the $q_i$'s are effectively one-hot encoded, where the only $q_i=1$ is that of the consensus label. In order to calculate the CXE for the entire data set, we simply perform the sum in Equation \ref{CXE2}
\begin{equation}
	\mathrm{CXE} = \sum_{j=1}^{N} -\mathrm{log}(p_{consensus}^j)
	\label{CXE2}
\end{equation}
where $p_{consensus}^j$ is the model output distribution's probability of the consensus class for image $j$ (summed over $N$ images). Notice that all references to the $q_i$ are gone since the effective one-hot encoding renders them irrelevant.

The second loss function---Kullback-Leibler divergence \citep{KLD}---is defined in Equation \ref{KLD}.
\begin{equation}
	D_{KL}(p||q) = H_q(p) - H(p) % = - \sum_{i=0}^{N-1} q_i \mathrm{log}\bigg(\dfrac{p_i}{q_i}\bigg)
	\label{KLD}
\end{equation}
This function is simply the difference between the model output, $p$, with respect to the target distribution, $q$, and the Shannon entropy of the model output, $p$. A major difference between this function and the sparse CXE is that the target distribution is no longer one-hot encoded. Instead, we use the NDA. Also, as with the CXE, one calculates the KLD for the entire dataset by summing $D_{KL}(p||q)$ for the individual images.

The main motivation for using the NDA instead of one-hot encoding on the consensus is to be able to utilize more information in the dataset which was being discarded by the one-hot encoding. Although this does introduce annotator disagreement into the loss function, it also introduces information about the similarity of different characters. This will allow for more avenues for future analysis on character misclassifications which are due to the characters having similar shapes. Furthermore, as will be shown in the next section, adoption of the NDA also causes the KLD-ResNet outputs to tend to have higher entropy than those of the CXE-ResNet.

\subsection{Ensemble Model: Stacked Generalization and $k$-Nearest Neighbors}

As will be shown below, though both ResNets performed well, there is a sizable subset of the data which one network classified correctly while the other was incorrect. Because of this disagreement between the ResNets, it is possible to construct an ensemble model which uses the two models' outputs to achieve an accuracy greater than that of the two individual networks, as informed by \citet{deepL}. We employ the technique of ``stacked generalization'', outlined in \citet{stackGen}, whereby the output distributions of the two ResNets are used as inputs to a new model. Using two ResNets with different loss functions decreases the amount of correlation between the models, giving us a more robust ensemble model. We use a $k$-nearest neighbors model (hereafter KNN) with $k=50$. We create the inputs for the KNN in the following manner. Let $C^i_j, K^i_j$ be the probability (taken from the CXE- and KLD-ResNet outputs, respectively) that the $i$-th image belongs to the $j$-th class. We can then create a new set of input data $X$ for the KNN model via Equation \ref{concat}.
\begin{equation}
    \begin{split}
        X^{1,\cdots,N}_{1,\cdots,M} & = C^{1,\cdots,N}_{1,\cdots,M} \\
	    X^{1,\cdots,N}_{M+1,\cdots,2M} & = K^{1,\cdots,N}_{1,\cdots,M}
    \end{split}
    \label{concat}
\end{equation}
Note that the lower index of $X$ takes on $2M$ values since there are two models each with $M=24$ classes whose output probabilities must be incorporated.

Similar to the ResNets, the KNN also produces an output distribution of classification probabilities which is based on the fraction of neighbors of different classes. This allows us to perform the same entropic analyses on all three models.

\subsection{Evaluation Metrics}

While the CXE-ResNet utilizes the standard accuracy metric for evaluation, the approach used by the KLD-ResNet is atypical of deep learning approaches. This is by nature of the labeling scheme used. Thus, Mean-Absolute-Error (MAE) was utilized as the primary metric. This posed a challenge when comparing the results from training the two models. The most straight-forward way of comparing accuracy for our models is to run inference on the entire dataset, including both training and validation data, and calculating the accuracy of model predictions. For consistency, we continue this approach in analysis of the ensemble model.

\subsection{Measuring Classification Uncertainty with Shannon Entropy}

Techniques exist to minimize and measure classification uncertainty using varied approaches. As \citet{Murtaza2020} explains, combining multiple annotations can often reduce misclassifications. Other approaches, such as \citet{Hendrycks2017}, seek to utilize the magnitude of Softmax maxima as a misclassification indicator. Yet another approach, in \citet{Granese2021},  sought to make use of Softmax distributions which yielded a 4\% decrease in misclassifications. We take an approach that has much in common with past efforts and utilizes the Shannon entropy of the ResNets' output probabilities, as in \citet{entropy1997}. This approach is well suited for our work because of the probabilistic nature of the NDA, which is atypical of labeling schemes, allowing us to compare the uncertainties of the human annotations and ResNet output distributions.

\section{Numerical Experiments}
We now move on to discuss the different numerical experiments performed on the data. We will begin with a discussion of the experimental setup and then proceed to compare the results of the different models tested. We will also provide some analysis of these results based on the Shannon entropy of the models' output distribution. For brevity, we only share here a limited number of plots consisting of the results for the entire dataset and results for the single character Alpha ($A \alpha$). Hundreds of plots were produced showing results for all classes. These plots, the ResNet models, and a demonstrational Jupyter notebook are available in an online repository available at \textit{https://github.com/mis2n/JDMDH-NDA-Results} .

\subsection{Experimental setup}
For measuring the accuracy of our models, we make use of the precision and recall statistics, given in Equation \ref{precrec}
\begin{gather}
    \begin{split}
	\textrm{Precision} & = TP/(TP+FP) \\
	\textrm{Recall} & = TP/(TP+FN)
	\label{precrec}
    \end{split}
\end{gather}
where $TP$ is the number of true positives, $FP$ is the number of true negatives, and $FN$ is the number of false negatives.

\subsection{Results}
Table \ref{confCK} shows the number of correct and incorrect classifications made by the CXE- and KLD-ResNet models. While both models correctly classify approximately 93\% of the images (see below for a more thorough look at model accuracy), the off-diagonal elements in the table show that there is a sizeable fraction of images which one model classified correctly while the other classified incorrectly. As stated earlier, stacked generalization can provide a boost in accuracy in such a case, where there is enough disagreement between two well-performing models that the best of both can be incorporated.

\begin{figure}
    \begin{minipage}[c]{0.45\linewidth}
        \centering
        \includegraphics[width=\linewidth]{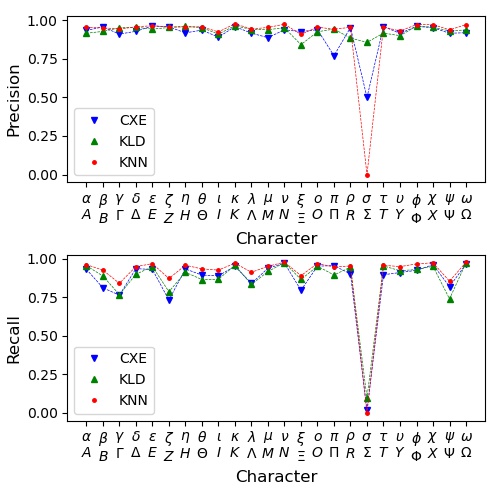}
	   \caption{Per character precision and recall for each of the three models. Note that     the KNN ensemble model is consistently higher than the individual models for most       characters.}
	   \label{accP}
    \end{minipage}
    \hfill
    \begin{minipage}[c]{0.45\linewidth}
        \centering
        \includegraphics[width=\linewidth]{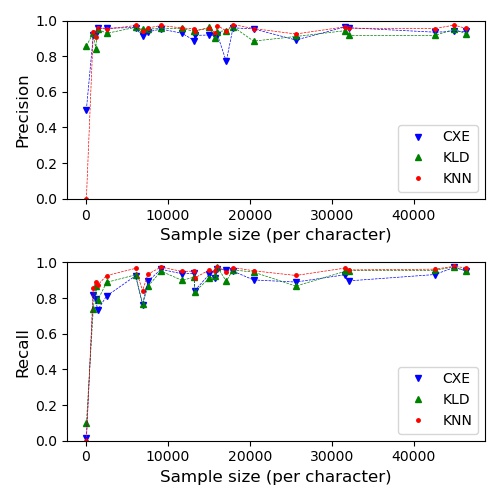}
	   \caption{Precision and recall of all three models versus per character sample size.     Notice the drop in recall for characters with smaller sample sizes}
	   \label{PrecRecVsSamp}
    \end{minipage}
\end{figure}

Table \ref{accT} and Figure \ref{accP} contain the precision and recall for the CXE-ResNets, KLD-ResNets, and KNN ensemble models per character and for the entire data set. Overall, the models achieved accuracies of 92.6\%, 93.2\%, and 95.6\%, respectively.

\begin{table}
    \begin{minipage}[c]{0.45\linewidth}
	\centering
    \tiny
	\begin{tabular}{ || c || c c | c c | c c || }
		\hline
		 & \multicolumn{2}{| c |}{CXE} & \multicolumn{2}{| c |}{KLD} & \multicolumn{2}{| c ||}{KNN} \\
		 \cline{2-7}
							& \multicolumn{2}{| c |}{pre. \hspace{8pt} rec.}
							& \multicolumn{2}{| c |}{pre. \hspace{8pt} rec.}
							& \multicolumn{2}{| c |}{pre. \hspace{8pt} rec.} \\
		\hline
		\hline
		Total 				& \multicolumn{2}{| c |}{0.928 \hspace{0pt} 0.926}
							& \multicolumn{2}{| c |}{0.932 \hspace{0pt} 0.932}
							& \multicolumn{2}{| c ||}{0.956 \hspace{0pt} 0.956} \\
		\hline
		\hline
		$A \alpha$	& \multicolumn{2}{| c |}{0.935 \hspace{0pt} 0.932}
			& \multicolumn{2}{| c |}{0.916 \hspace{0pt} 0.956}
			& \multicolumn{2}{| c |}{0.954 \hspace{0pt} 0.962} \\

		$B \beta$	& \multicolumn{2}{| c |}{0.957 \hspace{0pt} 0.812}
			& \multicolumn{2}{| c |}{0.928 \hspace{0pt} 0.888}
			& \multicolumn{2}{| c |}{0.952 \hspace{0pt} 0.927} \\

		$\Gamma \gamma$	& \multicolumn{2}{| c |}{0.910 \hspace{0pt} 0.764}
			& \multicolumn{2}{| c |}{0.951 \hspace{0pt} 0.767}
			& \multicolumn{2}{| c |}{0.942 \hspace{0pt} 0.841} \\

		$\Delta \delta$	& \multicolumn{2}{| c |}{0.928 \hspace{0pt} 0.936}
			& \multicolumn{2}{| c |}{0.955 \hspace{0pt} 0.900}
			& \multicolumn{2}{| c |}{0.958 \hspace{0pt} 0.950} \\

		$E \epsilon$	& \multicolumn{2}{| c |}{0.964 \hspace{0pt} 0.931}
			& \multicolumn{2}{| c |}{0.943 \hspace{0pt} 0.950}
			& \multicolumn{2}{| c |}{0.964 \hspace{0pt} 0.968} \\

		$Z \zeta$	& \multicolumn{2}{| c |}{0.957 \hspace{0pt} 0.733}
			& \multicolumn{2}{| c |}{0.953 \hspace{0pt} 0.787}
			& \multicolumn{2}{| c |}{0.957 \hspace{0pt} 0.872} \\

		$H \eta$	& \multicolumn{2}{| c |}{0.918 \hspace{0pt} 0.934}
			& \multicolumn{2}{| c |}{0.962 \hspace{0pt} 0.913}
			& \multicolumn{2}{| c |}{0.956 \hspace{0pt} 0.958} \\

		$\Theta \theta$	& \multicolumn{2}{| c |}{0.936 \hspace{0pt} 0.894}
			& \multicolumn{2}{| c |}{0.952 \hspace{0pt} 0.866}
			& \multicolumn{2}{| c |}{0.958 \hspace{0pt} 0.934} \\

		$I \iota$	& \multicolumn{2}{| c |}{0.891 \hspace{0pt} 0.890}
			& \multicolumn{2}{| c |}{0.910 \hspace{0pt} 0.868}
			& \multicolumn{2}{| c |}{0.924 \hspace{0pt} 0.927} \\

		$K \kappa$	& \multicolumn{2}{| c |}{0.959 \hspace{0pt} 0.951}
			& \multicolumn{2}{| c |}{0.965 \hspace{0pt} 0.958}
			& \multicolumn{2}{| c |}{0.978 \hspace{0pt} 0.971} \\

		$\Lambda \lambda$	& \multicolumn{2}{| c |}{0.917 \hspace{0pt} 0.841}
			& \multicolumn{2}{| c |}{0.940 \hspace{0pt} 0.834}
			& \multicolumn{2}{| c |}{0.943 \hspace{0pt} 0.913} \\

		$M \mu$	& \multicolumn{2}{| c |}{0.887 \hspace{0pt} 0.939}
			& \multicolumn{2}{| c |}{0.939 \hspace{0pt} 0.918}
			& \multicolumn{2}{| c |}{0.953 \hspace{0pt} 0.951} \\

		$N \nu$	& \multicolumn{2}{| c |}{0.938 \hspace{0pt} 0.972}
			& \multicolumn{2}{| c |}{0.951 \hspace{0pt} 0.975}
			& \multicolumn{2}{| c |}{0.974 \hspace{0pt} 0.978} \\

		$\Xi \xi$	& \multicolumn{2}{| c |}{0.926 \hspace{0pt} 0.796}
			& \multicolumn{2}{| c |}{0.842 \hspace{0pt} 0.866}
			& \multicolumn{2}{| c |}{0.909 \hspace{0pt} 0.891} \\

		$O o$	& \multicolumn{2}{| c |}{0.939 \hspace{0pt} 0.952}
			& \multicolumn{2}{| c |}{0.925 \hspace{0pt} 0.952}
			& \multicolumn{2}{| c |}{0.956 \hspace{0pt} 0.969} \\

		$\Pi \pi$	& \multicolumn{2}{| c |}{0.770 \hspace{0pt} 0.956}
			& \multicolumn{2}{| c |}{0.943 \hspace{0pt} 0.895}
			& \multicolumn{2}{| c |}{0.942 \hspace{0pt} 0.947} \\

		$P \rho$	& \multicolumn{2}{| c |}{0.953 \hspace{0pt} 0.902}
			& \multicolumn{2}{| c |}{0.884 \hspace{0pt} 0.946}
			& \multicolumn{2}{| c |}{0.955 \hspace{0pt} 0.953} \\

		$\Sigma \sigma$	& \multicolumn{2}{| c |}{0.500 \hspace{0pt} 0.016}
			& \multicolumn{2}{| c |}{0.857 \hspace{0pt} 0.097}
			& \multicolumn{2}{| c |}{0.000 \hspace{0pt} 0.000} \\

		$T \tau$	& \multicolumn{2}{| c |}{0.959 \hspace{0pt} 0.897}
			& \multicolumn{2}{| c |}{0.917 \hspace{0pt} 0.954}
			& \multicolumn{2}{| c |}{0.957 \hspace{0pt} 0.958} \\

		$\Upsilon \upsilon$	& \multicolumn{2}{| c |}{0.915 \hspace{0pt} 0.911}
			& \multicolumn{2}{| c |}{0.900 \hspace{0pt} 0.922}
			& \multicolumn{2}{| c |}{0.930 \hspace{0pt} 0.949} \\

		$\Phi \phi$	& \multicolumn{2}{| c |}{0.960 \hspace{0pt} 0.925}
			& \multicolumn{2}{| c |}{0.962 \hspace{0pt} 0.931}
			& \multicolumn{2}{| c |}{0.972 \hspace{0pt} 0.967} \\

		$X \chi$	& \multicolumn{2}{| c |}{0.952 \hspace{0pt} 0.962}
			& \multicolumn{2}{| c |}{0.955 \hspace{0pt} 0.951}
			& \multicolumn{2}{| c |}{0.972 \hspace{0pt} 0.974} \\

		$\Psi \psi$	& \multicolumn{2}{| c |}{0.916 \hspace{0pt} 0.820}
			& \multicolumn{2}{| c |}{0.930 \hspace{0pt} 0.740}
			& \multicolumn{2}{| c |}{0.938 \hspace{0pt} 0.855} \\

		$\Omega \omega$	& \multicolumn{2}{| c |}{0.919 \hspace{0pt} 0.966}
			& \multicolumn{2}{| c |}{0.937 \hspace{0pt} 0.974}
			& \multicolumn{2}{| c |}{0.972 \hspace{0pt} 0.977} \\
		\hline
	\end{tabular}
	\caption{Overall and per-character precision and recall (with respect to the human consensus) for the CXE, KLD, and KNN ensemble models.}
	\label{accT}
    \end{minipage}
    \hfill
    \begin{minipage}[c]{0.45\linewidth}
        \centering
	   \includegraphics[width=\linewidth]{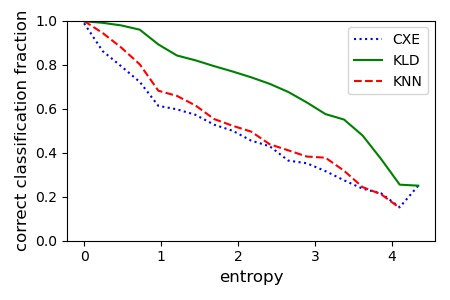}
	       \captionof{figure}{Plots of the fraction of correct classifications versus entropy of the output distribution for each model. Note that the KLD model is \textbf{NOT} better than the other models. Rather, it has consistently higher entropy values than the other models.}
	   \label{FracCorrect}
    \end{minipage}
\end{table}

Note that for the majority of characters and models, the precision and recall are $>$0.9, with the KNN tending to have the highest values. Cases with lower values for these statistics tend to be characters with fewer sample images (see Figure \ref{PrecRecVsSamp}). The most notable of these would be Sigma ($\Sigma \sigma$), for which there are only 62 samples ( compared to $>$42,000 for Alpha ($A \alpha$) ). While the two ResNets had precisions of 0.500 and 0.857 for Sigma ($\Sigma \sigma$), respectively, the recalls where $<$0.1 for both. Even more significant is the fact that the KNN achieved both precision and recall of 0.0. This is reasonable since the number of neighbors (50) is near the number of samples (62), making the chance of a majority of Sigma ($\Sigma \sigma$) neighbors quite low.

\subsection{Entropy Analysis}

\begin{figure}
    \begin{minipage}[c]{0.45\linewidth}
        \centering
        \includegraphics[width=\linewidth]{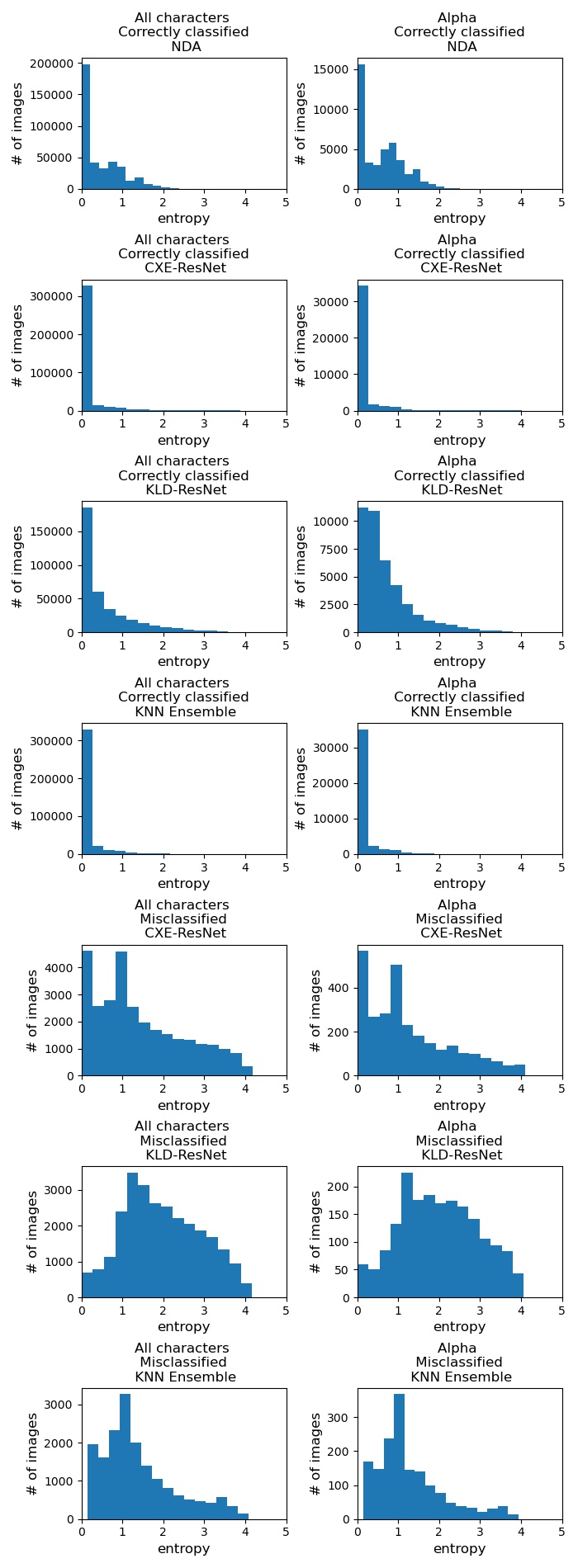}
	       \caption{Histogram of entropies for the NDAs and the three models' output distributions for both the entire data set and solely the character Alpha ($A \alpha$). We split the histograms into two sets: 1) images whose model classification agreed with the human consensus and 2) those which disagreed.}
	   \label{EntHist}
    \end{minipage}
    \hfill
    \begin{minipage}[c]{0.45\linewidth}
        \centering
        \includegraphics[width=\linewidth]{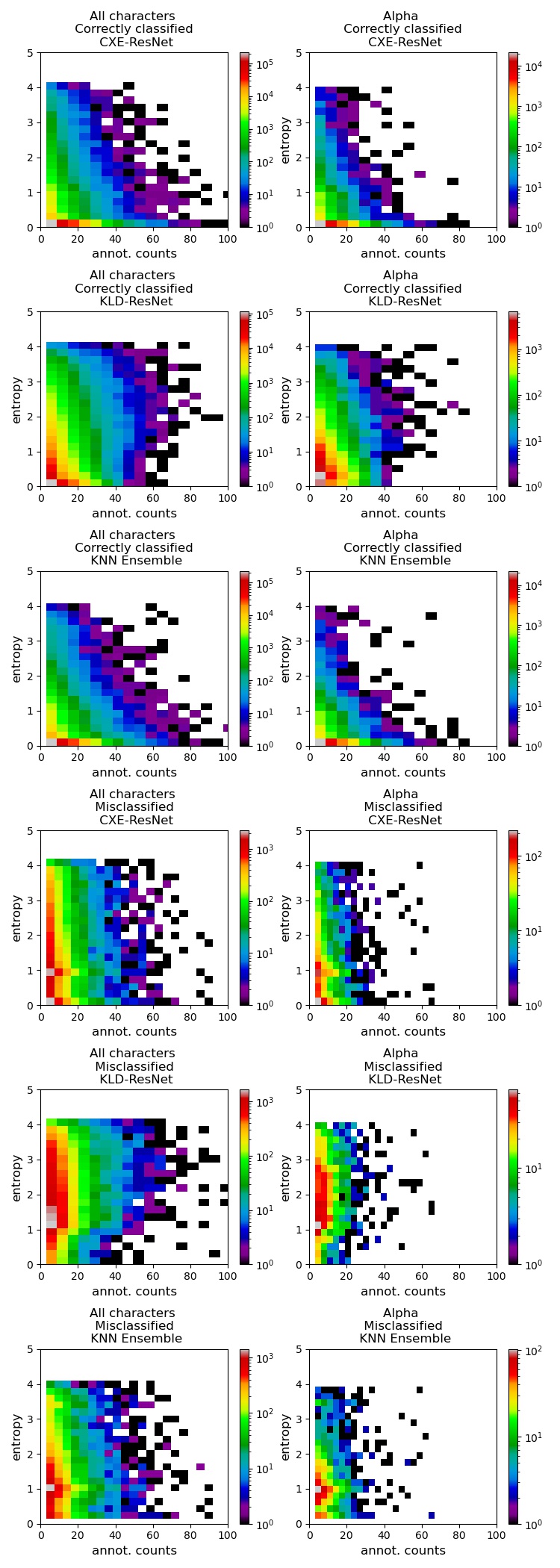}
            \caption{Two-dimensional histograms of the three models' output entropies versus the number of human annotations. The left column contains all characters and the right column contains only Alpha ($A \alpha$). We further split the plots into two portions: 1) images whose model classification agreed with the human consensus and 2) those which disagreed. Notice how the mean entropy value tends to decrease for large annotator counts for the correctly classified images.}
            \label{EntVsAnnot}
    \end{minipage}
\end{figure}

We also performed entropic analyses on the NDA, ResNets, and ensemble models' output distributions. Figure \ref{EntHist} shows histograms for the entropy of each of these distributions for both the entire data set and for just the character Alpha. The plots are split into two, with one plot containing only those images which were correctly classified by the respective model and the other plot containing those which were misclassified.

There are several interesting points revealed by these plots. First---and perhaps most readily noticed---is the fact that the majority of images have a low entropy in their respective output distributions, regardless of which model was used. For the NDA and correctly classified model histograms, the first bin has the greatest number of images by far. This means that for most images which are classified correctly, the model tends to be quite certain in its classification. For the misclassified model histograms, however, this is not the case (with the exception that the CXE-ResNet histograms in the first bin is approximately equal to the fourth). In fact, the tails of these histograms tend to contain the majority of samples. This means that for images which are classified incorrectly, the models tended to have less certainty in classification. Because of this discrepancy between the entropies of correct and incorrect classifications, entropy can serve not only as a good measure of classification uncertainty, but also as a predictor of classification accuracy (we discuss this further below). Images which have lower entropy will tend to be classified correctly while images with higher entropy will tend to be classified incorrectly. Figure \ref{FracCorrect} shows the fraction of correctly classified images versus entropy for the three models, illustrating the previous point well. Note, however, that the plot should not be interpreted as the KLD-ResNet significantly outperforming the other models. Rather, by nature of its labeling scheme, its output distributions have a high entropy on average.

Second, note the difference in Figure \ref{EntHist} between the three models' entropy profiles. The KLD-ResNet tends to have higher entropy values than both the CXE-ResNet and ensemble model. The discrepancy in entropy between the CXE-ResNet and KLD-ResNet is reasonable, since the CXE-ResNet only took into account the consensus label while the KLD-ResNet used the entire NDA. Additionally, it is reasonable for the ensemble entropies to be quite small since a KNN model's output distribution is merely derived from the fractions of neighbors of each class. Due to the clustering of the CXE-ResNet and KLD-ResNet output distributions (the input features for the ensemble model), the majority of samples of a given character are surrounded in feature space by other samples of the same character, inducing a low entropy in the KNN's output distribution.

Figure \ref{EntVsAnnot} shows two-dimensional histograms of the entropy of the models' output distributions for each image against the number of human annotations each image received.
The most notable feature of these plots is related to the difference in entropy profiles between the correctly classified and misclassified images. For the correctly classified images, we see that the range and mean of entropy values decreases as the annotation count increases. On the other hand, for the misclassified images, we see that the range and mean of entropy values does not tend to decrease (or even vary strongly) as annotator count increases. The CXE and KNN plots appear to show a very slight instance of this trend while the KLD plot seems to show the opposite (with the mean entropy increasing as annotation count increases).

These trends are reasonable when considered in conjunction with the law of large numbers. One might imagine in the limit of an infinite number of annotations, each image would have an NDA which converges to some distribution, say, $P$. Consequently, the entropy $H(P)$ would also converge. Also, as discussed above, images which are classified correctly tend to have low entropies. Therefore, as the annotation count for a given image (which would be correctly classified) increases, the entropy will tend to converge to a small value (though likely non-zero). Concerning plots of the misclassified data, we do not see a consistent trend as we did with the correctly classified data. For the KLD-ResNet plots, we see that the entropy tends to take on higher values at higher annotation counts. For the CXE-ResNet and KNN plots, it is difficult to discern a clear trend.

\subsubsection{Entropy as a predictor for classification accuracy}

As mentioned above, the entropy of a given image's output distribution (regardless of which model is used) is effective for predicting whether that image will be classified correctly by the model. To illustrate this, we trained a Support Vector Machine (SVM) to take the entropy of a model's output distribution for a given image as a single input feature and predict whether the model which produced this distribution would correctly classify the respective image. Thus, there are only two classes: 1) images correctly classified by the model or 2) images misclassified by the model. 

\begin{figure}[!htb]
	\centering
	\includegraphics[width=\linewidth]{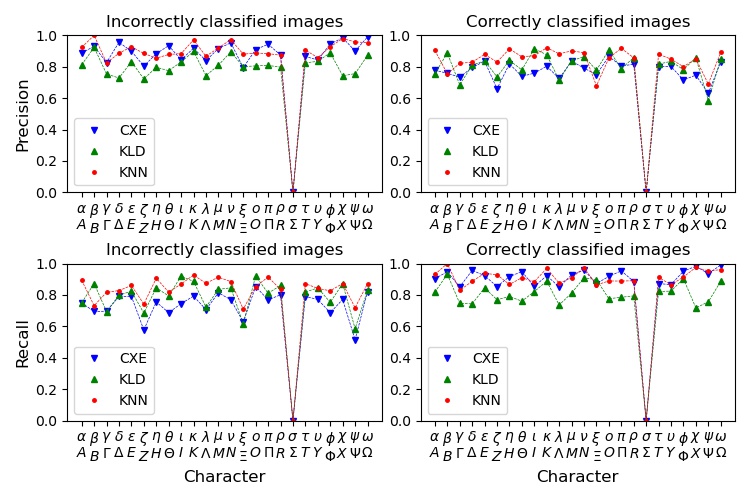}
	\caption{Per character precision and recall of the SVM model which predicts model classification accuracy based on the entropy of Softmax outputs.}
	\label{EntPred}
\end{figure}

We trained a separate SVM for each character, since each character has a different entropy profile and a different number of samples. We also used an 80/20 train/test split for each character. Additionally, due to the massive imbalance in the number of correctly and incorrectly classified characters, we balanced these data by choosing an equal number of correctly and incorrectly classified samples. Figure \ref{EntPred} depicts the results of this experiment, providing precision and recall statistics for both classes for all models and characters. It should be noted that it is unnecessary to display the precision \textit{and} recall for both classes of a classification problem since there is a dependency relation between these statistics, but both are displayed for clarity. 

For the CXE-ResNet, precision and recall values for both classes and all characters (other than Sigma) were between 0.486 and 1.000. For the KLD-ResNet, they were between 0.523 and 0.937. For the KNN, they were between 0.636 and 0.975. It should be noted that these SVMs tend to be biased toward predicting that the ResNets and ensemble models will correctly classify images. When examining their confusion matrices, there was a consistently larger number of false positives (where the SVM incorrectly predicted that the ResNets or ensemble models' classification would be correct) than false negatives. Also, it can be seen from Figure \ref{EntPred} that there is a consistent trend across the different characters where both the recall of the incorrectly classified images (bottom left) and the precision of the correctly classified images (top right) are lower than their converse values.

\section{Conclusions}

We have demonstrated that the multi-valued NDA labeling scheme used in our KLD model performs comparably to the standard, single-value labeling used in the CXE model. The CXE and KLD models largely agree with one another, but not always. This ensures that the ensemble model can utilize the differences between the component models effectively. We have also shown that by performing stacked generalization, with our pair of well-performing ResNet models, we are able to increase our prediction accuracy to a level beyond that achieved by either of the original models. We have also shown that the Shannon entropy of our models' output distributions is both an excellent measure of classification uncertainty and a predictor of classification accuracy for our data, with misclassified images tending to have a higher entropy than those correctly classified.  Although our techniques have only been applied to a single crowdsourced dataset (AL-ALL), we believe it may be a useful approach for other datasets to reduce the effects of ground-truth uncertainty in labeling. The amalgamation of these results yield models with improved classification trustworthiness, as well as the means to quantify the trustworthiness of individual classification predictions based on the annotation distributions (NDA).

Future work may focus on the removal of ambiguous and mislabeled images from the AL-ALL dataset by manually inspecting images whose NDA shows high entropy. There are a significant number of images in the dataset which have both a high NDA entropy and a high annotation count. Images such as these are often unintelligible, so removing them from the dataset may be beneficial when training models. Conversely, there is a possibility that such noise may be useful in avoiding model-overfitting, similar to the effects of image augmentation. Also, there are many images which are clearly intelligible but have been misclassified by the human annotators. These may be identified by using the ensemble model's results as the ground-truth which the human annotators must predict. Images where the humans disagree with the model when the model is clearly correct (determined by visual inspection of the image) can either be corrected or removed.

\bibliographystyle{plainnat}
\bibliography{jdmdh-example}

\end{document}